\documentclass[letterpaper, 10 pt, conference]{ieeeconf}
\IEEEoverridecommandlockouts                              

\usepackage{soul} 

\usepackage[usenames, dvipsnames]{color} 
\usepackage{graphicx} 
\usepackage[cmex10]{amsmath} 
\usepackage{amsfonts}
\usepackage{verbatim} 
\usepackage{subfigure}
\usepackage{multirow}
\usepackage{cite}
\usepackage{hhline}
\usepackage{bm}
\usepackage{tabularx,booktabs}
  \newcolumntype{Y}{>{\centering\arraybackslash}X}
\usepackage[bottom]{footmisc}

\usepackage{algorithm}
\usepackage[noend]{algpseudocode}

\bstctlcite{IEEEexample:BSTcontrol}

\soulregister\cite7
\soulregister\ref7

\begin{document}
\title{\LARGE \bf Sensor Observability Index: Evaluating Sensor Alignment for Task-Space Observability in Robotic Manipulators} 

\author{Christopher~Yee~Wong and Wael~Suleiman
\thanks{This work was supported in part by the Fonds de Recherche du Qu\'{e}bec - Nature et technologies. (\emph{Corresponding author: Christopher Yee Wong.})}
\thanks{C. Y. Wong and W. Suleiman are with the Universit\'{e} de Sherbooke, Sherbrooke, Canada (e-mail: \texttt{christopher.wong2}, \texttt{wael.suleiman (at) usherbrooke.ca}).}
}

\maketitle
\thispagestyle{empty} 
\pagestyle{empty} 

\begin{abstract}
In this paper, we propose a preliminary definition and analysis of the novel concept of sensor observability index.
The goal is to analyse and evaluate the performance of distributed directional or axial-based sensors to observe specific axes in task space as a function of joint configuration in serial robot manipulators.
For example, joint torque sensors are often used in serial robot manipulators and assumed to be perfectly capable of estimating end effector forces, but certain joint configurations may cause one or more task-space axes to be unobservable as a result of how the joint torque sensors are aligned.
The proposed sensor observability provides a method to analyse the quality of the current robot configuration to observe the task space.
Parallels are drawn between sensor observability and the traditional kinematic Jacobian for the particular case of joint torque sensors in serial robot manipulators.
Although similar information can be retrieved from kinematic analysis of the Jacobian transpose in serial manipulators, sensor observability is shown to be more generalizable in terms of analysing non-joint-mounted sensors and other sensor types. 
In addition, null-space analysis of the Jacobian transpose is susceptible to false observability singularities.
Simulations and experiments using the robot Baxter demonstrate the importance of maintaining proper sensor observability in physical interactions.
\end{abstract}

%
%

\section{Introduction} \label{sec:intro}
Sensors are invaluable tools for robots, as they are their way to observe themselves (introspection) and the world around them (extrospection), even in ways beyond the capability of humans. 
Unfortunately, sensors have limitations beyond their technical specifications, particularly those that are directional. 
Directional sensors are those with explicit axes along which measurements are performed, e.g. joint torque sensors, strain gauges, accelerometers, gyroscopes, distance sensors, etc. 
It is often a common assumption that, given the presence of sensors, all quantities are fully observable at all times.
For example, for a serial robotic manipulator with joint torque sensors at each joint, one would normally assume that the end effector (EE) forces could be reconstituted from the joint torque sensors \cite{VanDamme2011ICRA-EEforceestimationJTS, Phong2012ICRA-EEforceestimationJTS} for use in compliant control either directly \cite{Wu1980-ManipComplianceJointTorqueControl} or through machine learning \cite{Berger2020ICHR-LearningForceEstimation}.
One can hypothesize, and this paper demonstrates it, that this assumption is not always true. 
It is in fact possible that particular robot configurations lead to cases where the joint torque sensors are unable to observe certain forces at the end effector.
Another potential case where this assumption may not be true is a robot equipped with an array of distributed distance sensors on the arms, if the sensors were sparsely placed \cite{Stavridis2020ICHR-MobileManipDistributedDistanceSensors}.
Certain robot configurations may lead to potentially unobserved directions.

If a robot unknowingly enters a configuration where observations along certain axes can no longer be made, the end result could be disastrous for the robot, the task, or the environment.
These situations must be avoided in critical applications such as physical human-robot interaction and minimally invasive surgery. 
While the use of multi-axis sensors, e.g. 6-axis force-torque sensors, may observe all axes and prevent this issue, they may not always be available for many lower-cost and simpler systems given their cost, and alternative methods must be used \cite{Hawley2019IJHR-ExtForceObsforSmallHumanoids}. 
Furthermore, interactions with the robot body may not be observed if the sensor is mounted on an unconstrained end effector.

\begin{figure}[t]
\centering
\subfigure[]{
	\includegraphics[width=0.37\textwidth]{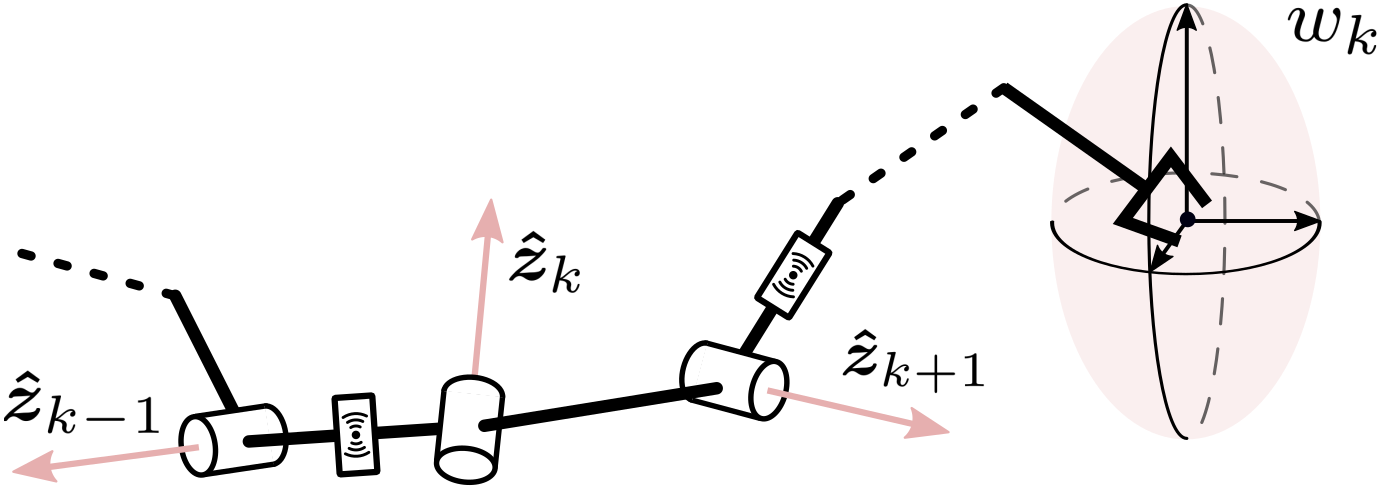}
	\label{fig:intro-kin}}
\subfigure[]{
	\includegraphics[width=0.39\textwidth]{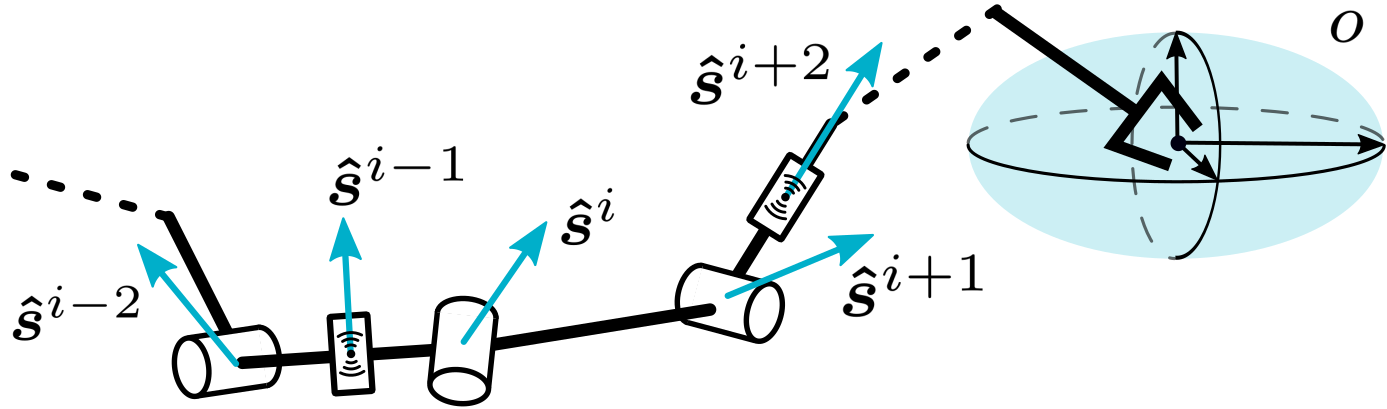}
	\label{fig:intro-sens}}
\caption{Comparison of a) joint axes $\bm{\hat{z}}_k$ and kinematic manipulability $w_k$ and b) positioning of joint-mounted and non-joint-mounted sensors $\bm{\hat{s}}^i$ and the sensor observability $o$ and their respective ellipsoids.}
\label{fig:smi-calc}
\end{figure}

Thus, a tool is required to analyse a particular robot configuration and provide a measure or index on the \textit{task-space observability} of the configuration.
Parallels can be drawn with the concept of traditional kinematic robot manipulability analysis \cite{Yoshikawa1985IJRR-Manipulability}, a quality measure of a robot's mobility and closeness to kinematic singularities.
In this paper, we extend this concept to task-space observability of robots based on the robot joint configuration and sensor placement.
As this issue of task-space sensor observability is highly dependent on the sensor configuration and kinematic structure of the robot, not all robots are equally affected.
Such an analysis is especially important for robots that do not have enough sensors to cover all task space dimensions reliably and must prioritize one over others.

%
%
%



\subsection{Background}
\label{sec:background}

Note that vectors and matrices are represented by bold-faced lower case and upper case letters, respectively, whereas scalar values are not bold-faced.
The traditional analytical Jacobian matrix $\bm{J(q)} \in \mathbb{R}^{{n_t} \times {n_q}}$ is defined as the matrix of first order partial derivatives relating ${n_q}$ joint-space velocities $\bm{\dot{q}}$ to ${n_t}$ task-space velocities $\bm{\dot{x}}$ \cite{Spong2020-RobotModelingandControl}:

\begin{equation}
\bm{\dot{x}} = \bm{J(q)}\bm{\dot{q}}, \ \ \bm{J(q)} = \begin{bmatrix}
  \frac{\partial x_1}{\partial q_1} & 
    \dots & 
    \frac{\partial x_1}{\partial q_{n_q}} \\[1ex] 
  \vdots & 
    \ddots & 
    \vdots \\[1ex]
  \frac{\partial x_{n_t}}{\partial q_1} & 
    \dots & 
    \frac{\partial x_{n_t}}{\partial q_{n_q}}
\end{bmatrix}
\label{eq:Jreg}
\end{equation}
In typical cases, the task space is defined as end effector position (${n_t} = 3$) or pose (${n_t} = 6$).
For readability, we will no longer explicitly write the Jacobian's dependency on the vector of joints $\bm{q}$, in other words $\bm{J(q)} \rightarrow \bm{J}$. 
The Jacobian can be used as a tool to measure different properties of the robot, notably to verify whether a specific robot configuration is at a \emph{kinematic singularity}.

The \emph{kinematic manipulability index} $w_k$, commonly referred to as simply the \emph{manipulability}, is a scalar quality measure of the robot's ability move in the task space based on the current joint configuration. 
Manipulability can also be used as a scalar measure of a robot's closeness to a kinematic singularity \cite{Yoshikawa1985IJRR-Manipulability}. 
It is an important tool that allows postures to be evaluated based on their mobility:

\begin{equation}
w_k = \sqrt{det(\bm{JJ}^T)}
\label{eq:manip_kin}
\end{equation}



\noindent $w_k$ can be exploited in different ways, typically with optimization algorithms to ensure that the robot motions stay away from any singularities \cite{Dufour2020JIRS-ManipulabilityQP}.
While the index $w_k$ is a scalar measure, the manipulability \emph{ellipsoid}, shown in Fig. \ref{fig:smi-calc} \cite{Yoshikawa1985IJRR-Manipulability, Chiu1987ICRA-ManipEllipsoids}, is a volumetric representation of a specific robot configuration's mobility that is proportional to the length of the ellipsoid principal axes.
Mobility, in this case, indicates the ease with which the end effector can move in a certain direction in task space proportional to joint motion.
As such, the manipulability ellipsoid can also be used as a target either as the main task or as a redundancy resolution sub-task \cite{Jaquier2018RSS-TrackingManipulabilityEllipsoids, Jaquier2021IJRR-GeometryAwareManipulabilityLearning}.
Controlling the manipulability ellipsoid ensures that a certain level of manipulability is present, especially if a particular shape is desired.

The concept of manipulability has greatly evolved to encompass different calculation methods and applications since its early introduction by Yoshikawa \cite{Yoshikawa1985IJRR-Manipulability}. 
For example, some authors modified the concept to instead calculate the manipulability of the centre of mass of floating base robots \cite{Gu2015ICRA-COMManipulabilityHumanoids, Azad2017ICRA-DynManipCOM}.
Manipulability has also been extended to multi-robot closed-chain systems \cite{BicchiTRA2000-ManipulabilityClosedChain} and continuum robots \cite{Gravagne2002TRA-ManipulabilityContinuum}.

\subsection{Contributions}
\label{sec:contributions}

In the same vein as the kinematic manipulability index and ellipsoid, we introduce the novel concept of the \emph{sensor observability index}, as well as the \emph{sensor observability ellipsoid}.
The proposed concepts qualitatively evaluate, based on the current joint configuration, the ability of distributed sensors to measure in the task space. 
While the analysis of distributed sensing is thoroughly studied in sensor networks and swarm robotics, the analysis introduced here is from the viewpoint of a single multi-jointed and articulated robot.
To the best of the authors' knowledge, the framework established here has not been proposed previously.
The paper first provides an analytical framework in Sec. \ref{sec:method} for defining task-space sensor observability based on the cumulative transformations of each individual sensor and draws analogies to the traditional Jacobian.
In Sec. \ref{sec:experiments}, we demonstrate the framework's utility by applying the analysis to a robot with joint torque estimation and its importance when resolving end effector forces. 
Finally, closing comments are provided in Sec. \ref{sec:concl}.

The derivations in this paper use force sensing as the case study for the observability of end effector forces \cite{VanDamme2011ICRA-EEforceestimationJTS, Phong2012ICRA-EEforceestimationJTS} as it is the most intuitive case.
The method can be applied to other types of axis-based sensors, but is not shown here for brevity. 
Furthermore, the proposed formulation also includes the possibility to analyse non-joint-mounted sensors, for example strain gauges, accelerometers, or distance sensors placed on a link, as shown in Fig. \ref{fig:intro-sens}.


\subsection{Differences versus State-Space Observability}
\label{sec:diff}

The terms \emph{observable} and \emph{observability} used in the context of sensor observability differ from that used in state-space representation.
State-space observability refers to \emph{state} observation---estimation of the internal system states based on the system output alone \cite{Spong2020-RobotModelingandControl}.
Instead, sensor observability only concerns the measurement of task-space quantities and not the measurement of system states nor does it require the full system dynamics to be known.
Further comparisons of two will be the subject of future work.

\section{Sensor Transformation and Observability Index} 
\label{sec:method}

Prior to introducing the concept of sensor observability, we would like to note that the analysis presented here assumes that joints do not have mechanical limits, and thus in this work we ignore special treatment of sensors at joint limits. 
Furthermore, we assume that sensors are bidirectional, in the sense that they are capable of measuring along both positive and negative directions of their sensing axis (for example, laser-based distance sensors are unidirectional, whereas accelerometers are bidirectional).
Unidirectional sensors require more complex sensor axis analyses and will be addressed in future work.
A summary of the method is presented in Algorithm \ref{alg:sens_obs}.

\begin{algorithm}[t]
\caption{Summary of Sensor Observability}
\begin{algorithmic}[1]
    \State $\bm{\hat{s}}^{\prime,i}$ for $i \in 1...n_s$  \Comment{Def. local sensor axes (Sec \ref{sec:localsensoraxis})}
    \State $\bm{\hat{s}}^i \leftarrow \bm{R}\bm{\hat{s}}^{\prime,i}$ \Comment{Rotate to match task frame}
		\State $\bm{\tilde{s}}^i \leftarrow T_\square(\bm{\hat{s}}^i)$ \Comment{Sensor-type transf. (Sec \ref{sec:sensortranform})}
		\State $\bm{S} = \begin{bmatrix} \bm{\tilde{s}}^1 \cdots \bm{\tilde{s}}^{n_s} \end{bmatrix}$  \Comment{Sens. obsv. matrix (Sec \ref{sec:sensobs})}
		\State $\bm{s} \leftarrow \Gamma_\square(\bm{S})$ \Comment{Sensor observability vector} 
		\State $o \leftarrow \prod_{j=1}^{n_t} \bm{s}_j$ \Comment{Sensor observability index}
		%
%
\end{algorithmic}
\label{alg:sens_obs}
\end{algorithm}

\subsection{Local Sensor Axis $\bm{\hat{s}}^i$}
\label{sec:localsensoraxis}
First, for each individually measured sensor axis $i \in 1...n_s$, we define a sensor axis vector $\bm{\hat{s}}^{\prime,i} \in \mathbb{R}^{n_t}$, where each element indicates whether a task-space axis is observed or not by taking on a value between $\{0,1\}$.
Note the difference between $n_s$, the number of sensor axes, and $n_t$, the number of task space coordinates.
A zero value means that that particular axis is not observed, whereas a value of one means that the axis is directly observed, i.e. the sensor axis is parallel with the task-space axis.
Values between 0 and 1 mean that the task space axis is only partially observed by an off-axis sensor.
The prime symbol in $\bm{\hat{s}}^{\prime,i}$ denotes that it is defined in the local $i$-th sensor frame $\mathcal{F}_i$.
A sensor axis vector $\bm{\hat{s}}^i$ without the prime symbol represents the set of axis vectors rotated to the task frame $\mathcal{F}_{EE}$.
For demonstration purposes, we set $\mathcal{F}_{EE}$ at the origin of the end effector, but aligned with the world frame.
For example, a single one-axis joint torque sensor could be seen as an element of $ SE(3)$ with $n_t = 6$ and represented by: 

\begin{equation}
\bm{\hat{s}}^{\prime}_{\tau z} = \begin{bmatrix} \bm{\hat{s}}_{p,{\tau z}}^{\prime} \\ \bm{\hat{s}}_{\theta, {\tau z}}^{\prime} \end{bmatrix} = \begin{bmatrix} 0 & 0 & 0 & 0 & 0 & 1 \end{bmatrix}^T
\label{eq:s_JTS}
\end{equation}

\noindent where $\bm{\hat{s}}^{\prime}_{\tau z}$ is in the local joint frame according to Denavit-Hartenberg (DH) parameters \cite{denavit1955TAM-DHparam} and $(\cdot)_p$ and $(\cdot)_\theta$ subscripts are the translational and rotational components, respectively. Similarly, a single-axis load cell in the $x$-axis is represented by $\bm{\hat{s}}^{\prime}_{fx} = \begin{bmatrix} 1 & 0 & 0 & 0 & 0 & 0 \end{bmatrix}^T$.
Multi-axis sensors, e.g. a 3-axis load cell that can detect forces in the $xyz$-axes but not torques, would be represented by the set of three individual sensor axis vectors, one in each $x$-, $y$-, and $z$-axis, \textit{i.e.} $\{\bm{\hat{s}}^{\prime}_{fx}, \bm{\hat{s}}^{\prime}_{fy}, \bm{\hat{s}}^{\prime}_{fz}\}$.
In the same vein, a 6-axis force-torque sensor would be the set of six individual sensor axes represented by $\{\bm{\hat{s}}^{\prime}_{fx}, \bm{\hat{s}}^{\prime}_{fy}, \bm{\hat{s}}^{\prime}_{fz}, \bm{\hat{s}}^{\prime}_{\tau x}, \bm{\hat{s}}^{\prime}_{\tau y}, \bm{\hat{s}}^{\prime}_{\tau z} \}$.

All local sensor axis vectors are then rotated to match the orientation of the task frame $\mathcal{F}_{EE}$, i.e. $\bm{\hat{s}}^i = \bm{R}\bm{\hat{s}}^{\prime,i}$.
It is important to note that the $n_s$ is defined as the number of individually measured sensor axes and not the number of physical sensors. 
For example, a robot with two physical 3-axis sensors would have $n_s = 6$, where sensor frames $\mathcal{F}_i \in i = \{1,2,3\}$ and $\mathcal{F}_i \in i = \{4,5,6\}$ are located at their respective physical sensors.
Defining $n_s$ in this manner simplifies the derivations that follow.

\subsection{Sensor Transformation $T_\square(\bm{\hat{s}}^i)$}
\label{sec:sensortranform}

The \emph{sensor transformation function} $T_\square(\bm{\hat{s}}^i)$ is defined as a sensor-type-dependent transformation that may be required when shifting frames.
For example, force-torque sensors may observe torques generated by linear forces at $\mathcal{F}_{EE}$ if there exists a moment arm, according to $\bm{\tau} = \bm{r} \times \bm{f}$. 
Thus, a force-torque sensor $\bm{\hat{s}}^i$ would undergo the following \emph{force} sensor transformation $T_f(\bm{\hat{s}}^i)$, designed by the subscript $f$, in the task frame:

\begin{equation}
\begin{aligned}
\bm{\tilde{s}}^i &= \begin{bmatrix} \bm{\tilde{s}}^i_{p} \\ \bm{\tilde{s}}^i_{\theta} \end{bmatrix} \\
&= T_{f}(\bm{\hat{s}}^i) = 
\begin{cases}
       \begin{bmatrix} \lvert\bm{\hat{s}}^i_{p}\rvert + \bm{0} \\ \lvert\bm{\hat{s}}^i_{\theta}\rvert \end{bmatrix} \text{,} &\text{if } \bm{r}^i \times \bm{\hat{s}}^i_{\theta} = \bm{0} \\
       \begin{bmatrix} \lvert\bm{\hat{s}}^i_{p}\rvert + \frac{\lvert\bm{r}^i \times \bm{\hat{s}}^i_{\theta}\rvert}{\|\bm{r}^i \times \bm{\hat{s}}^i_{\theta}\|}  \\ \lvert\bm{\hat{s}}^i_{\theta}\rvert \end{bmatrix} \text{,} &\text{otherwise.} \\ 
     \end{cases}
\label{eq:senstransf}
\end{aligned}
\end{equation}

\noindent where $\bm{r}^i$ is the position vector from the $i$-th sensor axis to the task frame $\mathcal{F}_{EE}$, $\lvert \cdot \rvert$ is the element-wise absolute function,
and $\|\cdot\|$ is the Euclidean norm to normalize the cross product as directional analysis of sensor axes should not be influenced by the magnitude of the moment arm.
The piece-wise defined function is used in the case where $\bm{r}^i$ and $\bm{\hat{s}}^i_{\theta}$ are collinear such that $\|\bm{r}^i \times \bm{\hat{s}}^i_{\theta}\| = 0$, which would result in an undefined fraction otherwise.
The result of this transformed sensor axis is designated by the tilde operator $\bm{\tilde{\cdot}}$, whereas previously the hat operator $\bm{\hat{\cdot}}$ designated a locally-defined sensor axis.

The use of the absolute function is two-fold: a) we assume that the sensors are bidirectional and b) it ensures that sensor axes do not subtract from each other.
It is important to note that the exact transformation $T_\square(\cdot)$ is dependent on the sensor type and the laws of physics that govern it. 
Certain transformation functions may simply be the identity function.

The method to interpret the transformed quantity $\bm{\tilde{s}}^i$ is as follows: each element of $\bm{\tilde{s}}^i$ represents a task-space axis that is observed by the various locally-defined terms of $\bm{\hat{s}}^i$ it contains.
For example, in (\ref{eq:senstransf}), given that both $\bm{\hat{s}}^i_{p}$ and $\bm{\hat{s}}^i_{\theta}$ terms appear in the linear force term $\bm{\tilde{s}}^i_{p}$, any linear forces at the EE would be observed by both the linear and rotational axes of the $i$-th sensor (if they exist).


\subsection{Sensor Observability Matrix $\bm{S}$, Vector $\bm{s}$, Function $\Gamma_\square$, Index $o$ and Ellipsoid}
\label{sec:sensobs}

We define the \emph{sensor observability matrix} $\bm{S} \in \mathbb{R}^{{n_t} \times {n_s}}$ as the matrix of column vectors of the transformed sensor axis vectors $\bm{\tilde{s}}^i$:

\begin{equation}
\bm{S} = \begin{bmatrix} \bm{\tilde{s}}^1 \cdots \bm{\tilde{s}}^{n_s} \end{bmatrix}
\label{eq:sensmatrix}
\end{equation}

Next, we define the \emph{sensor observability function} $\Gamma_\square(\bm{S})$ and the resultant \emph{sensor observability vector} $\bm{s} = \Gamma_\square(\bm{S})$ where $\bm{s} \in \mathbb{R}^{n_t}$ as the cumulative sensing capabilities of all individual sensors at the task frame $\mathcal{F}_{EE}$ with $n_t$ task axes.
The purpose of function $\Gamma_\square(\bm{S})$ is to synthesize the transformed sensor axes according to the desired metric for analysis. 
Here, we give two example definitions of $\Gamma_\square(\bm{S})$:

\subsubsection{Row-wise sum function}
\begin{equation}
\bm{s} = \Gamma_{sum}(\bm{S}) = \sum_{i=1}^{n_s} \bm{\tilde{s}}^i
\label{eq:obsfunc-sum}
\end{equation}


\subsubsection{Row-wise max function}
\begin{equation}
\bm{s} = \Gamma_{max}\left(\bm{S}\right) \text{ where } \forall j \in 1...{n_t} : s_j = \max_{i=1...n_s} \tilde{s}^i_j
\label{eq:obsfunc-max}
\end{equation}

\noindent where the subscript $j$ in $s_j$ and $\bm{\tilde{s}}_j$ indicates the $j$-th task-space axis of $\bm{s}$ and $\bm{\tilde{s}}$, respectively\footnote{Recall: superscript $i$ is for the $i$-th sensor axis, which is different from the subscript $j$ for the $j$-th task space axis (and similarly for subscript $k$, defined later).}, and also corresponds to the $j$-th row of $\bm{S}$.
The sum function $\Gamma_{sum}(\cdot)$, as the name implies, performs an row-wise summation across all transformed sensor axes in $\bm{S}$ and measures the cumulative task-space sensing capabilities across all sensors.
The summation can potentially provide a measure of redundancy if multiple sensors measure the same task space axis.
One potential issue with this method is that for the same value, the sum function does not differentiate between an axis that is directly observed by one or a few sensors, or only minimally observed by many off-axis sensors.
The lack of this differentiation may result in unintended low quality readings from non-closely aligned sensors. 

Conversely, the element-wise max function $\Gamma_{max}(\cdot)$ determines the maximum alignment between the individual sensor axes and each task space axis.
It provides a quality measure of how \emph{directly} a task-space axis is observed and, in a sense, its trustworthiness.
The max function $\Gamma_{max}(\cdot)$ is always constrained between $\{0,1\}$, where a value $s_j = 1$ indicates that there is at least one sensor that is directly and fully observing the $j$-th task space axis, while $s_j < 1$ indicates that it is only measured indirectly by all sensors.
In all cases, $s_j \approx 0$ would indicate that the $j$-th axis is in danger of no longer being observed. 

Other sensor observability functions may be used as well, depending on the preferred analysis.
For example, a sum with minimum thresholding or a squared sum could potentially negate the masking effect if many low quality observations by minimally observed sensor axes are present. 
Note that certain formulations of $\Gamma_\square(\cdot)$ may also use the sensor positions (in addition to the sensor orientations in $\bm{S}$) in case it is relevant, e.g. for modelling sensor-to-sensor interactions.
An example is discussed in Sec. \ref{sec:SOI-Advantages}.

Next, we define the \emph{sensor observability index} $o$: 

\begin{equation}
o = \prod_{j = 1}^{n_t} s_j
\label{eq:obsindex}
\end{equation}

Analogous to the kinematic manipulability index $w_k$ in (\ref{eq:manip_kin}), the sensor observability index $o$ is a scalar quality measure of task-space observability.
If any $s_j \rightarrow 0$, then $o \rightarrow 0$, and the system is at risk of being unable to observe one or more task space axes. 
The case where $o = 0$ is called an \emph{observability singularity} where the robot is in an \emph{observability singular configuration}, and the system has lost the ability to observe one or more task space axes.
This situation should be avoided for risk of potentially causing failure resulting from the robot being blind in certain task space axes.
As such, $o$ can be used as an optimization variable during motion planning to avoid low quality joint configurations.
While numerical interpretation of the sensor observability index is system-dependent, it can easily used as a relative gauge of system observability performance.

As sensor observability is a continuous scalar quality measure, it is difficult to pinpoint an exact threshold where one can say that observability has been lost.
An ideal sensor with infinite sensitivity and zero noise would be able to provide a usable reading for all $s_j > 0$ or $o > 0$. 
Realistically, a poorly observed axis is more susceptible to sensor noise $\epsilon$. 
Thus, sensor observability could potentially be flagged as lost when the signal noise is greater than the required minimum detectable amount (e.g. a system must be able to detect interaction forces of at least 10 N) as a function of the sensor sensitivity, i.e. when $s_j < f(\epsilon)$.
In addition, external factors can affect the observability threshold, e.g. large magnetic fields can interfere with inertial measurement units (IMU).
The effect of external factors could be measured and factored into calculating the minimum required observability threshold.
Observability threshold formulation will be the subject of future work. 

\begin{figure}[t]
\centering
\subfigure[]{
	\includegraphics[width=0.38\textwidth]{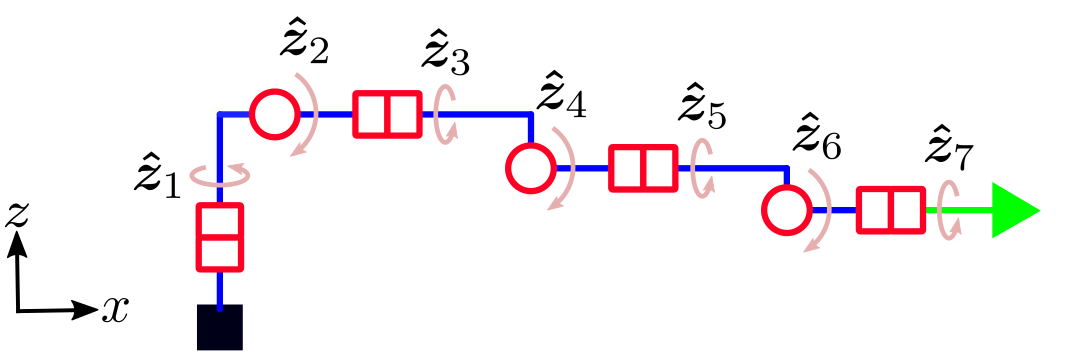}
	\label{fig:robotconfig-structure}}
\subfigure[]{
	\includegraphics[height = 4.75 cm]{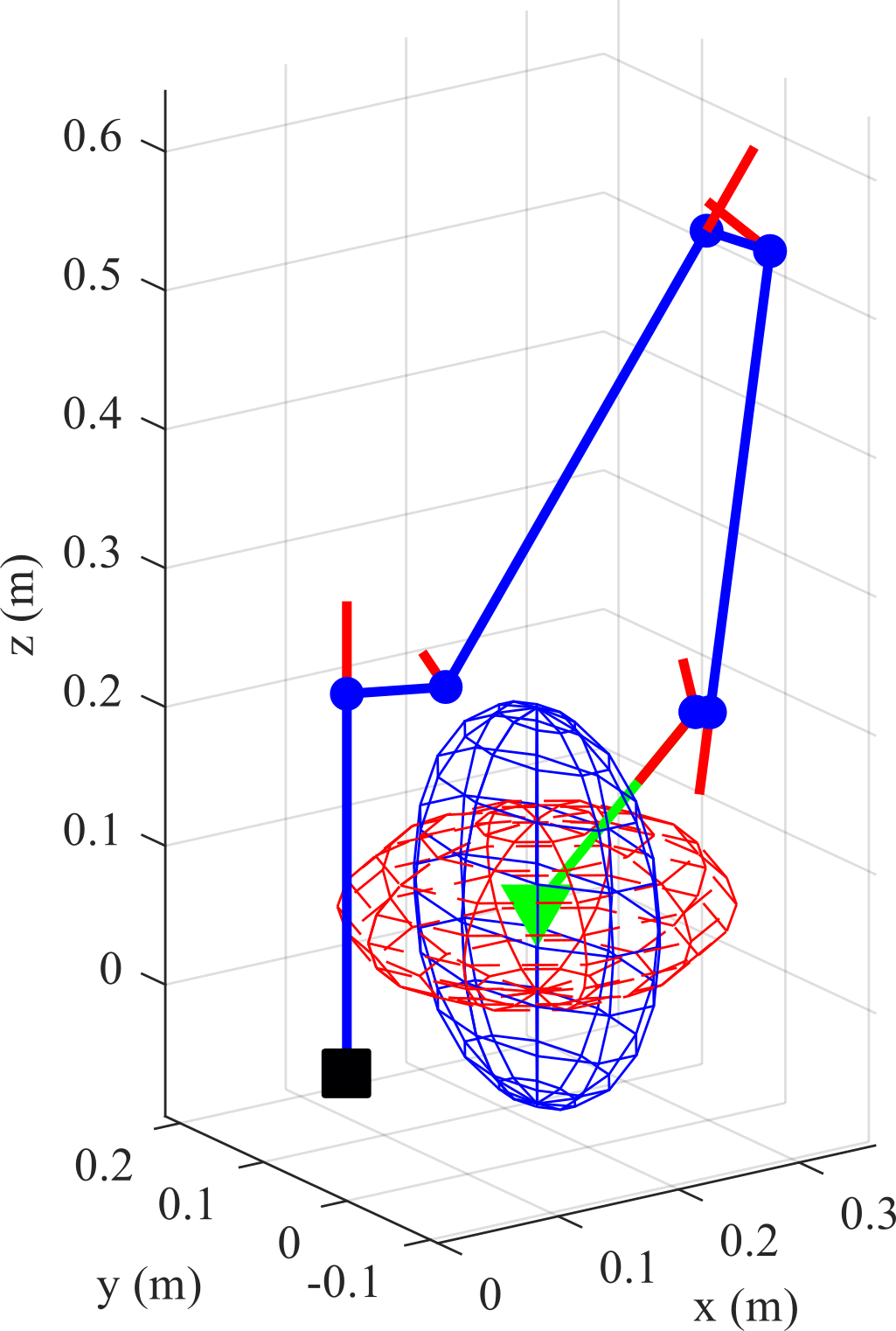}
	\label{fig:robotconfig-normal}}
\subfigure[]{
	\includegraphics[height = 4.75 cm]{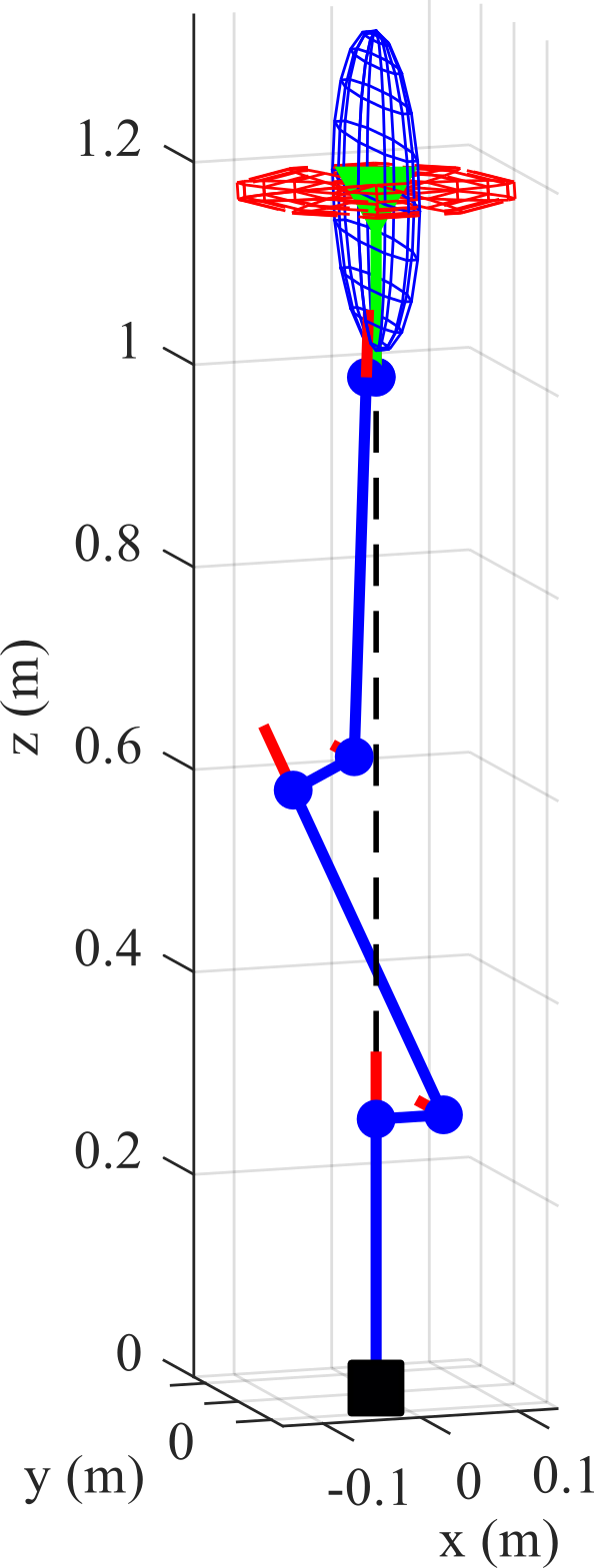}
	\label{fig:robotconfig-kinematicsingularity}}
\subfigure[]{
	\includegraphics[height = 4.75 cm]{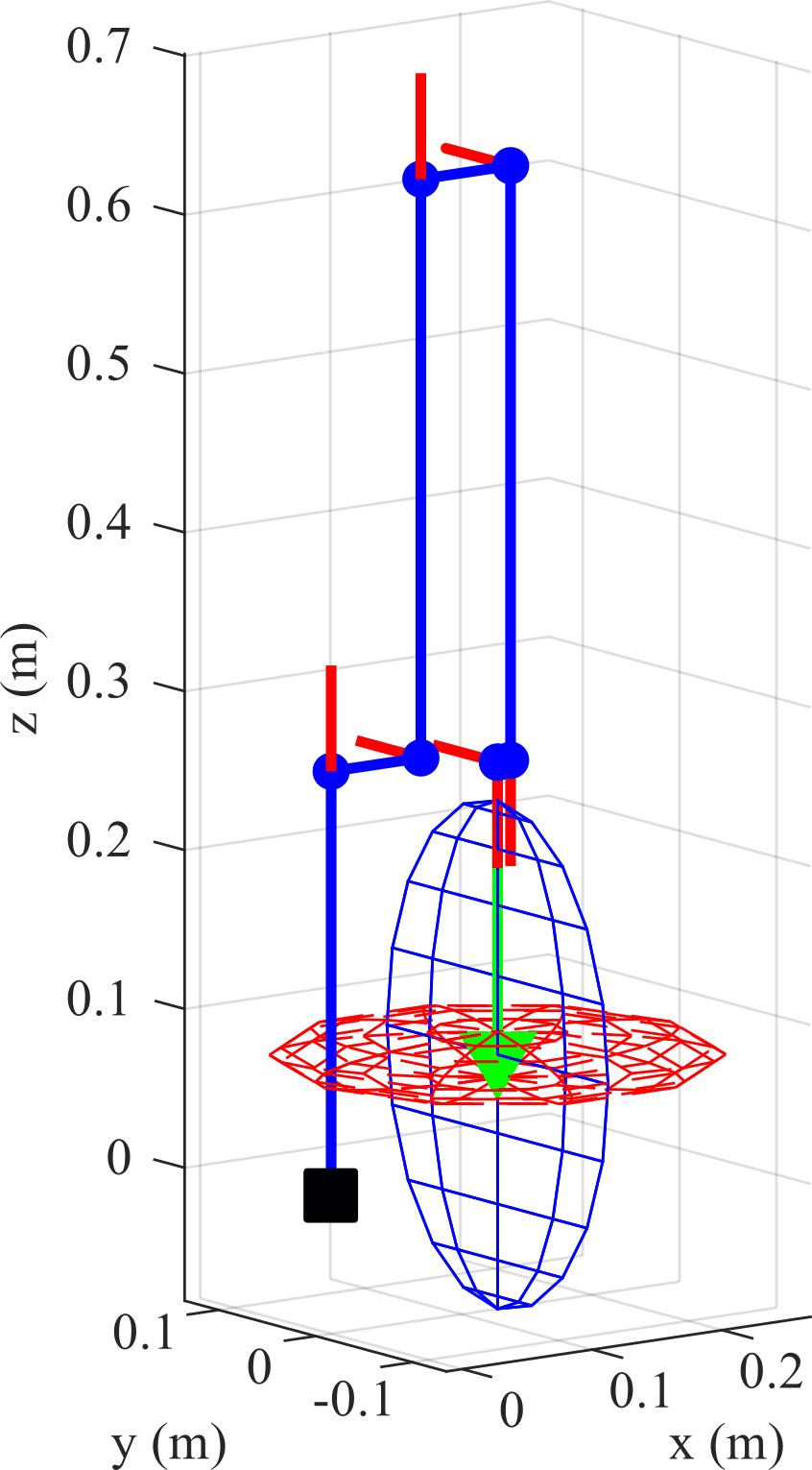}
	\label{fig:robotconfig-singularity}}
\subfigure[]{
	\includegraphics[width=0.35\textwidth]{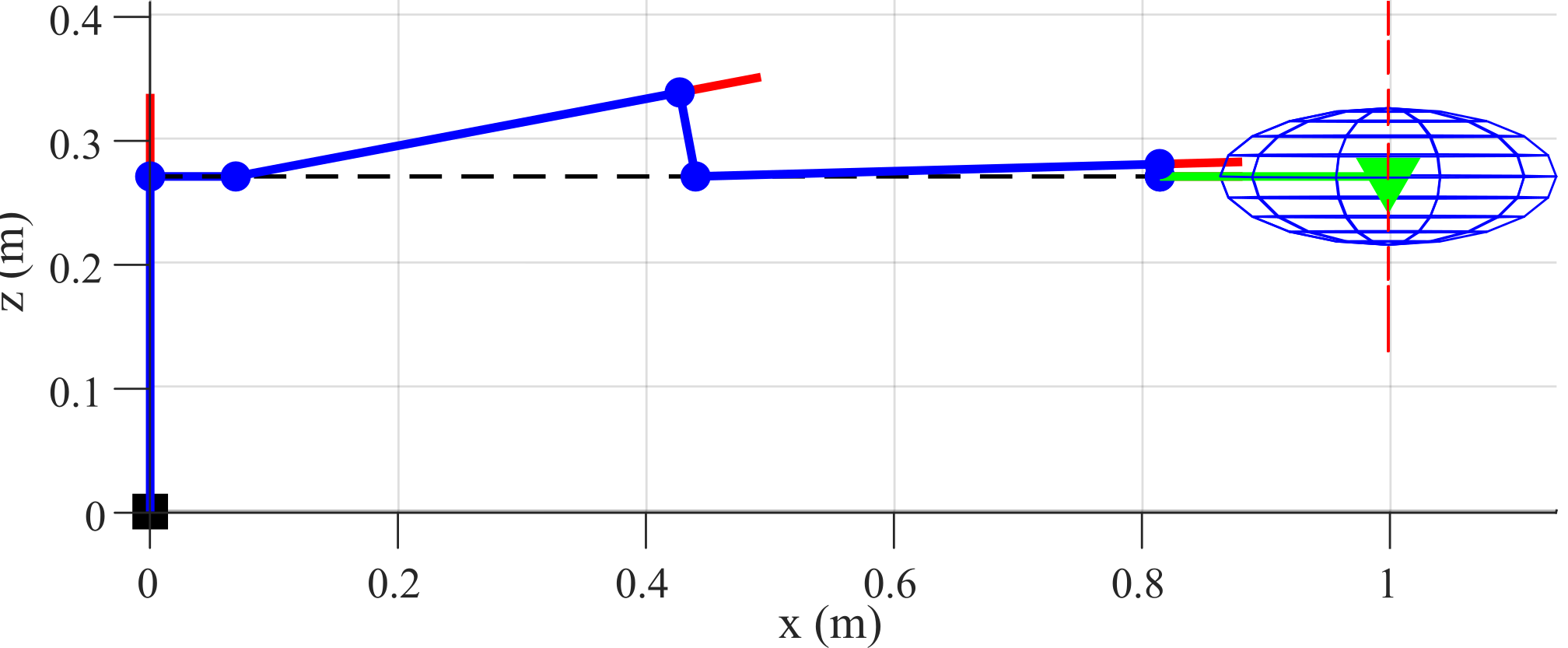}
	\label{fig:robotconfig-singsensX}}
\caption{a) Representation of a single 7-DOF Baxter robot arm with only traditional joint torque sensors in b) arbitrary configuration, c) kinematic singularity in $\theta_z$, d-e) sensor observability singularities in d) $\tau_x$ and e) $f_x$. Blue lines represent the robot links, red lines are joint axes, and the green triangle is the EE. Black square indicates arm base. 
Force and torque observability ellipsoids based on the sum function $\Gamma_{sum}(\cdot)$ are shown in red dashed and blue solid ellipsoids, respectively. 
Note that the ellipsoids in c) are very thin, but not completely flat, i.e. $o \neq 0$}
\label{fig:robotconfig}
\end{figure}

Similar to the manipulability ellipsoid defined previously in Sec. \ref{sec:background}, we define the sensor observability ellipsoid in $\mathbb{R}^{n_t}$ where the principal axes are proportional to the magnitude of the task-space observability.
Fig. \ref{fig:robotconfig} showcases various robot configurations and their resulting sensor observability ellipsoid.
For visualization purposes, sensor observability is split into the force $\bm{s}_{p}$ (red dashed line ellipsoid) and torque $\bm{s}_{\theta}$ (blue solid line ellipsoid) components.
In the arbitrary configuration shown in Fig. \ref{fig:robotconfig-normal}, all axes are observable, as can be seen by the 3D shape of the force and torque ellipsoids. 
In the observability singular configurations shown in Figs. \ref{fig:robotconfig-singularity} and \ref{fig:robotconfig-singsensX}, an axis of the observability ellipsoid collapses to zero. 
This indicates that the corresponding axis, torques along the $x$-axis in Fig. \ref{fig:robotconfig-singularity} and forces along the $x$-axis in Fig. \ref{fig:robotconfig-singsensX}, is not observable by the on-board joint torque sensors. 

\subsection{Special Case: Similarities to Kinematic Analysis} 
\label{sec:parallels}

In the special case where the sensor axes are collinear with the joint axes, parallels can be drawn between the sensor observability and kinematic analyses.
Let us examine a serial manipulator with only revolute joints and single-axis joint torque sensors located at each joint that are aligned with the joint axes.
In this specific case, each local sensor axis vector $\bm{\hat{s}}^{\prime,i} = \bm{\hat{s}}^{\prime}_{\tau z}$ as in (\ref{eq:s_JTS}). 
Thus, using the force sensor transformation $T_f(\bm{\hat{s}}^i)$ in (\ref{eq:senstransf}) and the sum-based observability function $\Gamma_{sum}(\bm{S})$ in (\ref{eq:obsfunc-sum}), we can see the that final sensor observability $\bm{s}$ is of form: 

\begin{equation}
\bm{s} = \sum_{i=1}^{n_s} \begin{bmatrix} \frac{\lvert\bm{r}^i \times \bm{\hat{s}}^i_{\theta}\rvert}{\|\bm{r}^i \times \bm{\hat{s}}^i_{\theta}\|} \\ \lvert\bm{\hat{s}}^i_{\theta}\rvert \end{bmatrix}
\label{eq:kinequiv-sens1}
\end{equation}

The term $\bm{\hat{s}}^i_{p}$ is absent from (\ref{eq:kinequiv-sens1}) as it is zero for single-axis joint torque sensors. The summation in (\ref{eq:kinequiv-sens1}) can be rewritten in matrix form using the sensor observability matrix $\bm{S}$ and a vector of ones $\bm{1}$:

\begin{equation}
\bm{s} = \begin{bmatrix} 
\frac{\lvert\bm{r}^1 \times \bm{\hat{s}}^1_{\theta}\rvert}{\|\bm{r}^1 \times \bm{\hat{s}}^1_{\theta}\|} & \dots & \frac{\lvert\bm{r}^{n_s} \times \bm{\hat{s}}^{n_s}_{\theta}\rvert}{\|\bm{r}^{n_s} \times \bm{\hat{s}}^{n_s}_{\theta}\|} \\
\lvert\bm{\hat{s}}^1_{\theta}\rvert & \dots & \lvert\bm{\hat{s}}^{n_s}_{\theta}\rvert 
\end{bmatrix}
\begin{bmatrix} 1 \\ \vdots \\ 1 \end{bmatrix}_{n_s \times 1} = \bm{S}\bm{1}
\label{eq:kinequiv-sens2}
\end{equation}

For the kinematic analysis, we begin with the geometric velocity analysis \cite{Spong2020-RobotModelingandControl}:

\begin{equation}
\begin{bmatrix} \bm{v} \\ \bm{\omega} \end{bmatrix} = \sum_{k = 1}^{n_q} \begin{bmatrix} \dot{q}_k\bm{\hat{z}}_k \times \bm{r}_k \\ \dot{q}_k\bm{\hat{z}}_k \end{bmatrix}
\label{eq:kinequiv-kin1}
\end{equation}

\noindent where $\bm{v}$ and $\bm{\omega}$ are the translational and angular velocities of the end effector, $\dot{q}_k$ is the angular velocity of the $k$-th joint, and $\bm{\hat{z}}_k$ is the $k$-th joint axis.
Similar to (\ref{eq:kinequiv-sens1})-(\ref{eq:kinequiv-sens2}), (\ref{eq:kinequiv-kin1}) may be rewritten in matrix multiplication form using the kinematic Jacobian $\bm{J}$ and the vector of joint angular velocities $\bm{\dot{q}}$:

\begin{equation}
\begin{bmatrix} \bm{v} \\ \bm{\omega} \end{bmatrix} = \begin{bmatrix} \bm{\hat{z}}_1 \times \bm{r}_1 & \dots & \bm{\hat{z}}_{n_q} \times \bm{r}_{n_q} \\
 \bm{\hat{z}}_1 & \dots & \bm{\hat{z}}_{n_q} \end{bmatrix} \begin{bmatrix} \dot{q}_1 \\ \vdots \\ \dot{q}_{n_q} \end{bmatrix} = \bm{J}\bm{\dot{q}}
\label{eq:kinequiv-kin2}
\end{equation}

Given that the joint torque sensors and joint axes are unit vectors and collinear, we in fact have $\bm{\hat{s}}^i_{\theta} = \bm{\hat{z}}_k$, $\bm{r}^i = \bm{r}_{k}$, and $n_q = n_s$.
Thus, equations (\ref{eq:kinequiv-sens2}) and (\ref{eq:kinequiv-kin2}) have very similar form despite differences in normalization where $\bm{S} \approx \bm{J}$ for a standard serial manipulator with joint torque sensors on each rotational joint.
To understand this relationship, joint axes could potentially be thought of as velocity measurement sensors.
A similar analysis holds for prismatic joints paired with single axis load cell.

Despite the similarity in form of (\ref{eq:kinequiv-sens2}) and (\ref{eq:kinequiv-kin2}), it is important to note that a sensor observability singularity does not necessarily imply kinematic singularity and vice versa.
For example, the configuration shown in Fig. \ref{fig:robotconfig-kinematicsingularity} is a kinematically singular configuration where axes 1 and 7 are collinear and $w_k = 0$, but is not an observability singularity $o \neq 0$ (the red ellipsoid is very thin but not completely flat).
Conversely, the joint configurations shown in Figs. \ref{fig:robotconfig-singularity} and \ref{fig:robotconfig-singsensX} are both observability and kinematic singularities, demonstrating potential overlaps between the two indices for this particular robot and sensor configuration.

\begin{figure}[t]
\centering
\includegraphics[width=0.37\textwidth]{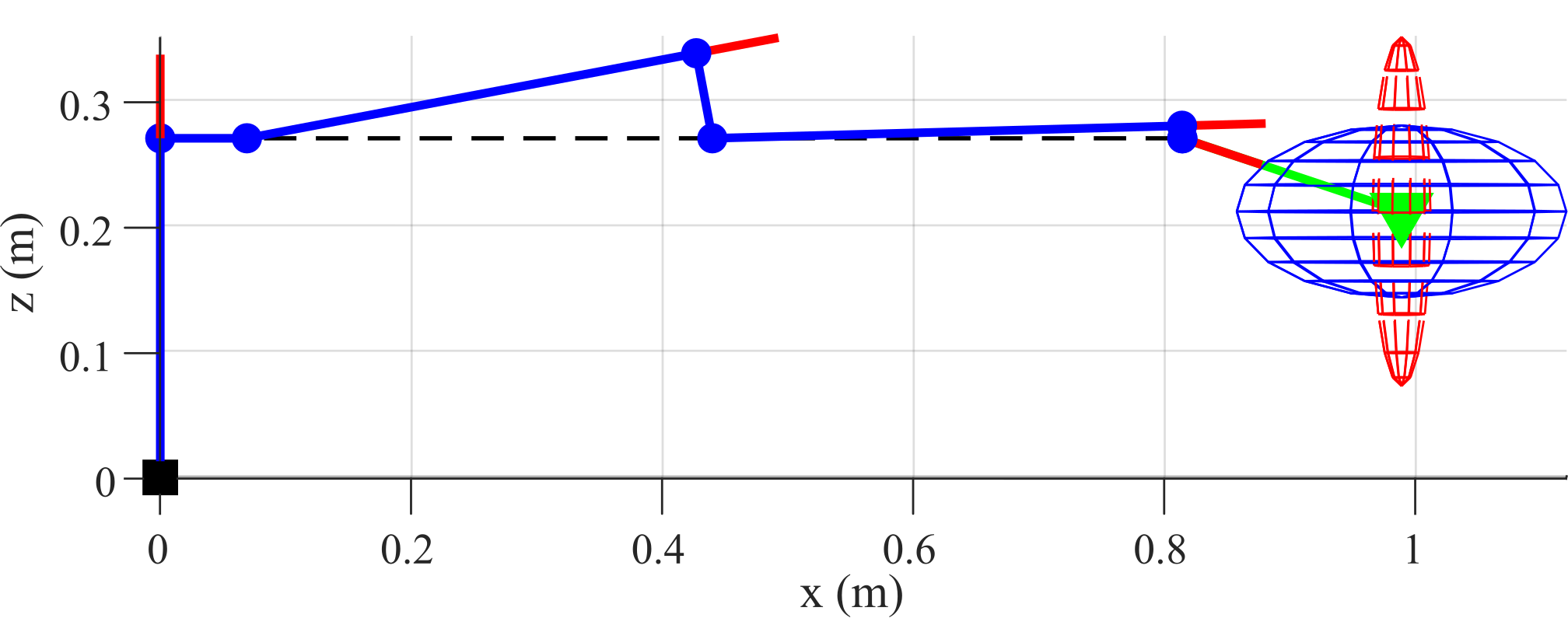}
\caption{Robot configuration where a non-zero null space vector exists for $\bm{J}^T$, but it is not a sensor observability singularity as $o \neq 0$. 
This configuration is similar to Fig. \ref{fig:robotconfig-singsensX} except that the end effector is slightly tilted downwards at the 6th joint.
}
\label{fig:robotconfig-nullspace}
\end{figure}

In this special case for serial manipulator robots, it is sometimes possible to extract similar information using the end effector force and joint torque relationship $\bm{\tau} = \bm{J}^T \bm{f}$ and examining the null space of $\bm{J}^T$.
The existence of non-zero null space vectors $\bm{J}^T$ indicates the possibility of having zero joint torques despite non-zero end effector forces, but it can be caused by two distinct cases.
One case is a sensing deficiency in the same manner as sensor observability. 
The other case occurs when end-effector forces and torques balance each other out and result in zero readings at the joints.
For example, in the configuration shown in Fig. \ref{fig:robotconfig-nullspace}, a null space analysis of $\bm{J}^T$ indicates that a force applied at the end effector in the $x$-axis can be nullified with a balancing torque in the $y$-axis, which result in zero joint torques.
Despite the existence of a non-zero null space vector for $\bm{J}^T$ in this configuration, this pose is not an observability singularity as $o = 57.63$ (so $o \neq 0$).
Thus, even in the special case where the sensor axes are collinear with the joint axes, null space analysis of the Jacobian cannot replace sensor observability analysis.
The reason is due to the use of absolute values in (\ref{eq:senstransf}) and (\ref{eq:kinequiv-sens1}) during sensor observability analysis that negates the possibility of the sensor axes cancelling each other out.

\subsection{Advantages of Sensor Observability Analysis}
\label{sec:SOI-Advantages}


While the discussion above might lead one to think that sensor observability can simply be derived through the traditional kinematic analysis, the parallels drawn in Sec. \ref{sec:parallels} are applicable only in the special case where the sensor axes are collinear with the joint axes (e.g. standard serial manipulators, whether using revolute or prismatic joints). 
The advantage of sensor observability analysis is that it is flexible and applicable to different robot architectures and sensor mounting styles. 
Once different robot architectures are used, then the parallels with the traditional Jacobian formulation no longer apply.
For example, a robot using a load cell in the middle of a link could be a lower cost alternative to using a joint-mounted torque sensor.
In this case, since non-joint mounted sensors are present, the formulation of $\bm{S}$ will differ significantly from $\bm{J}$.
Similarly, the resulting indices $o$ and $w_k$ will differ even more.

Moreover, certain sensors may need to be interpreted differently than simply with axis direction, which is why the sensor transformation $T_\square(\bm{\hat{s}}^i)$ and sensor observability $\Gamma_\square(\bm{S})$ functions are implemented.
For example, an articulated robot may be covered with an array of distributed laser distance sensors \cite{Stavridis2020ICHR-MobileManipDistributedDistanceSensors}, ultrasonic sensors, or magnetic directional proximity sensors \cite{Wu2016AIM-MagneticProximitySensorSphericalRobot}.
Depending on the joint configuration, these sensors may interact with each other (e.g. ultrasonic interference or crossing magnetic fields), and these interactions may affect sensing quality.
Their interactions could thus be modelled and captured by the $T_\square(\bm{\hat{s}}^i)$ and $\Gamma_\square(\bm{S})$ functions and used to optimize joint configuration to minimize interference.

Thus, sensor observability analysis can be viewed as a more generalized and more flexible analysis than kinematic analysis using the Jacobian matrix.
Sensor observability analysis can also be used in the robot design phase to optimize the placement of sensors to either create redundancy or minimize the number of sensors required, and will be the subject of future work. 

\section{Results \& Discussion}
\label{sec:experiments}

\subsection{Simulation}
\label{sec:simulation}

\begin{figure}[t]
\centering
\subfigure[]{
	\includegraphics[width = 0.42\textwidth]{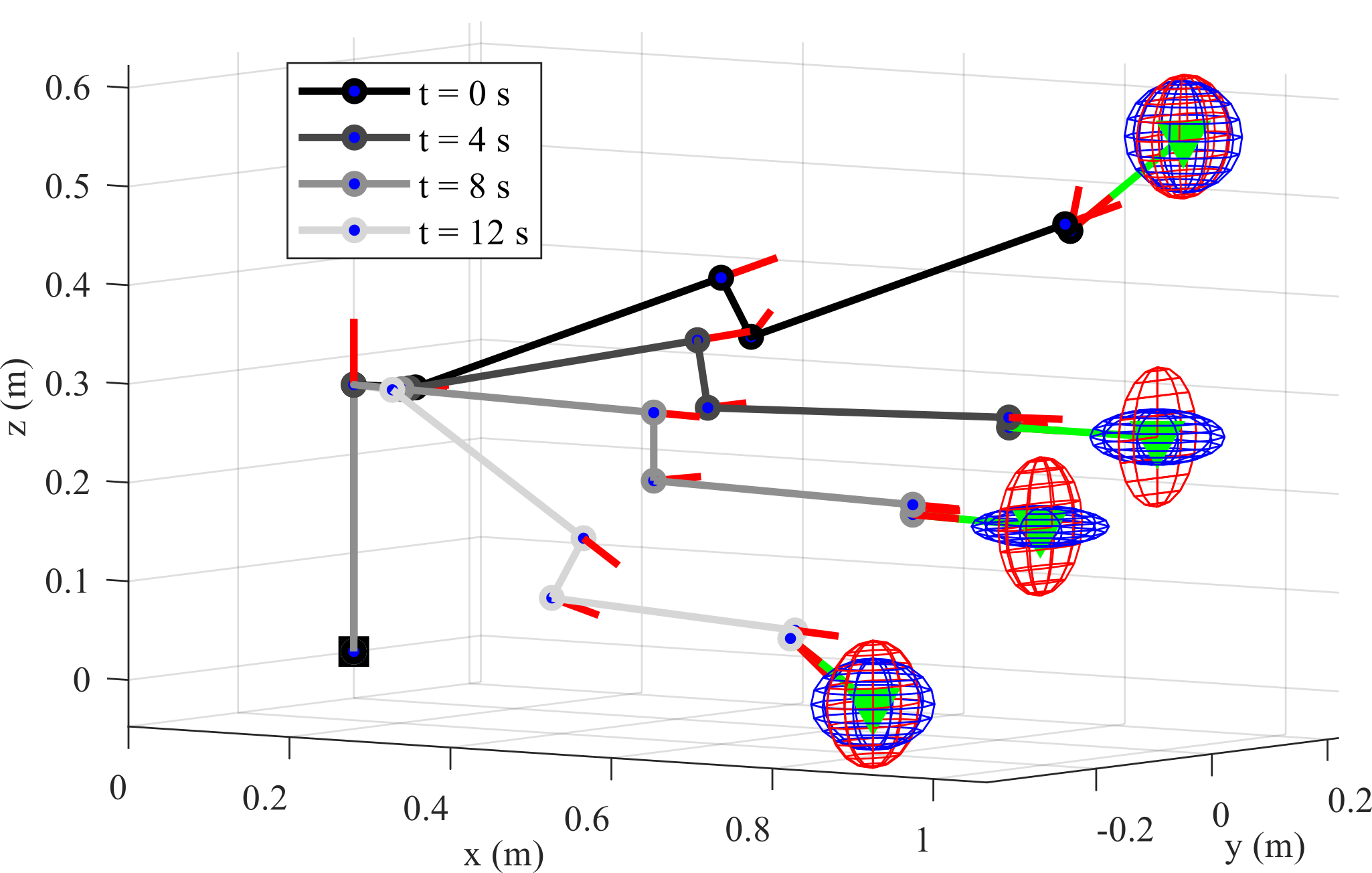}
	\label{fig:simsweep-robotconfig}}
\subfigure[]{
	\includegraphics[width = 0.43\textwidth]{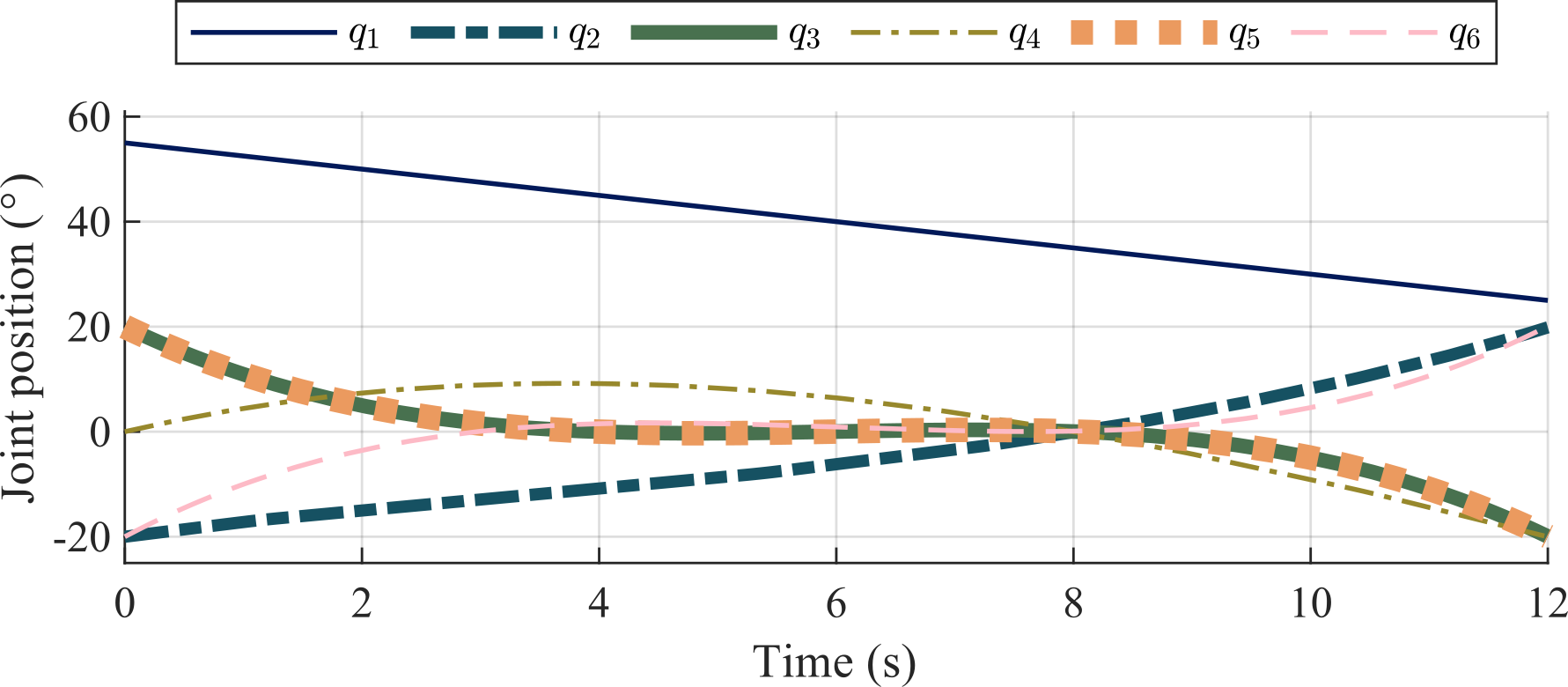}
	\label{fig:simsweep-jointpos}}
\subfigure[]{
	\includegraphics[width = 0.43\textwidth]{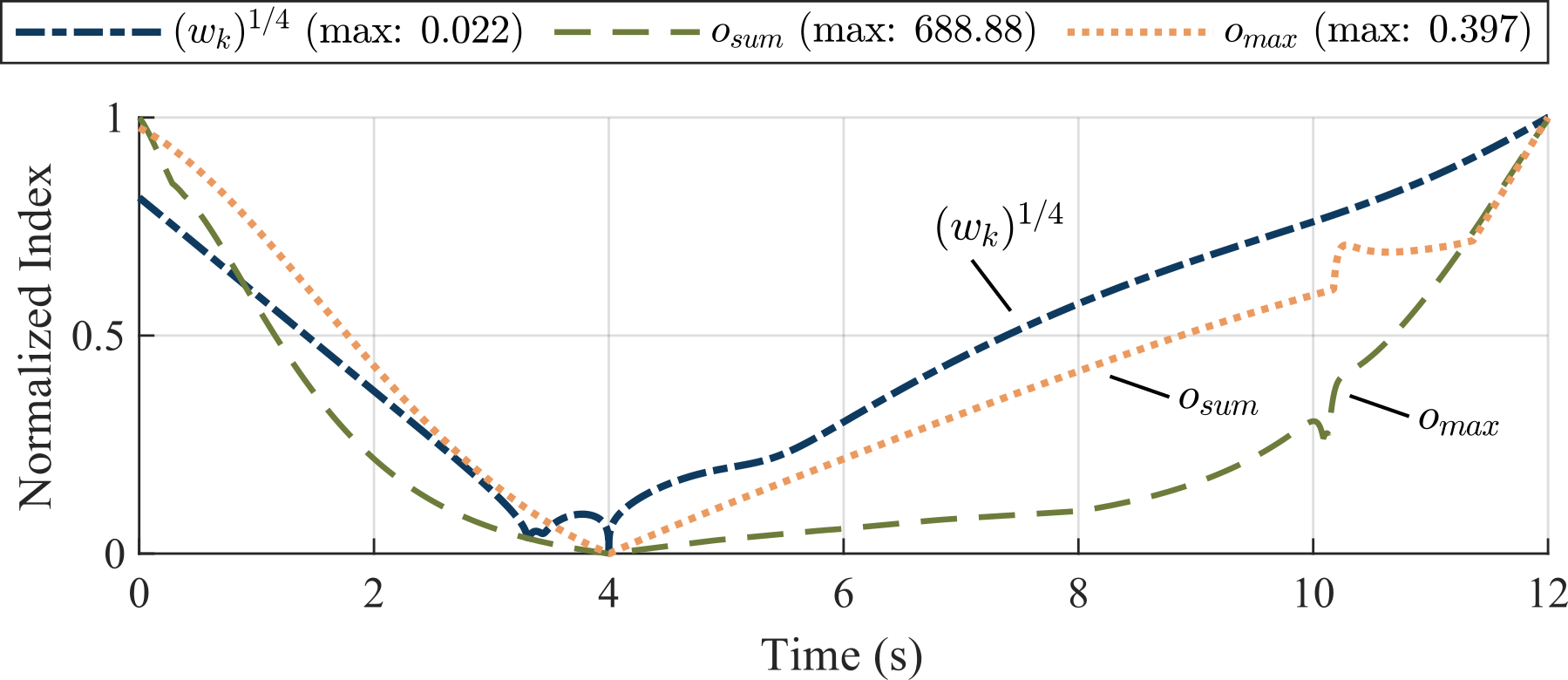}
	\label{fig:simsweep-index}}
\caption{a) Simulation of a single Baxter robot arm moving starting in an arbitrary position, sweeping through an observability singularity at $t = 4$ s (configuration shown in Fig. \ref{fig:robotconfig-singsensX}),
setting joints 2-6 $\bm{q}_{2-6} = 0$ at $t = 8$ s, 
and ending in an arbitrary position at $t = 12$ s. b) Joint positions of the maneuver. 
Joint $q_7$ is not shown as it is held at a constant $q_7 = 0$. 
c) Plot of the evolution of kinematic manipulability $w_k$ and 
observability index using the sum $o_{sum}$ and max $o_{max}$ functions. All indices are normalized to 1, but $w_k$ 
is further scaled with an exponential to emphasize the changes close to zero at $t = 4$ s.
}
\label{fig:simsweep}
\end{figure}

We use the fixed-base dual-arm robot Baxter from Rethink Robotics to demonstrate the importance of sensor observability analysis. 
Each Baxter arm contains 7 degrees of freedom whose kinematic structure is shown in Fig. \ref{fig:robotconfig-structure} and described in detail in \cite{Williams2017-BaxterKinematics}, and each joint contains a position encoder and the capability for torque estimation.
Given the structure of Baxter, there exists sensor observability singular configurations, as shown in Fig. \ref{fig:robotconfig}.

We simulate the kinematic structure of a single Baxter arm in MATLAB and sweep through multiple configurations shown in Fig. \ref{fig:simsweep-robotconfig}.
The robot begins in an arbitrary configuration at $t = 0$ s.
At $t = 4$ s, the robot moves to the configuration shown in Fig. \ref{fig:robotconfig-singsensX}, which incurs simultaneous sensor observability and kinematic manipulability singularities $o,w_k=0$.
At $t = 8$ s, joints 2-6 are set to zero, $\bm{q}_{2-6} = 0$.
The robot finally ends in another arbitrary configuration at $t = 12$ s.
The joint angles are plotted in Fig. \ref{fig:simsweep-jointpos}.
Joint $q_7$ is not shown in the plot as it is held at a constant $q_7 = 0$ and has no effect on the results.
Fig \ref{fig:simsweep-index} plots the evolution of the kinematic manipulability index $w_k$ and sensor observability index using the sum $o_{sum}$ in (\ref{eq:obsfunc-sum}) and max $o_{max}$ functions in (\ref{eq:obsfunc-max}) through the different robot configurations.
All indices are normalized, though $w_k$ is further scaled to emphasize its evolution particularly at $t = 4$ s.

As the robot moves towards the observability and kinematic singularity at $t = 4$ s, all indices approach zero.
This singular configuration eliminates the observability of force ($o_{sum}, o_{max} \rightarrow 0$) and translational motion ($w_k \rightarrow 0$) in the $x$-axis.
Although both $o_{sum}$ and $o_{max}$ hold somewhat similar trends, the max function $o_{max}$ has a maximum value of 1 whereas the index using sum function $o_{sum}$ is theoretically unbounded ($o_{sum}$ is normalized to the maximum of 688.88 in this scenario).




\subsection{Robot Interaction} 
\label{sec:exp-kinesteach}


\begin{figure}[!t]
\centering
\subfigure[]{
	\includegraphics[width = 0.225\textwidth]{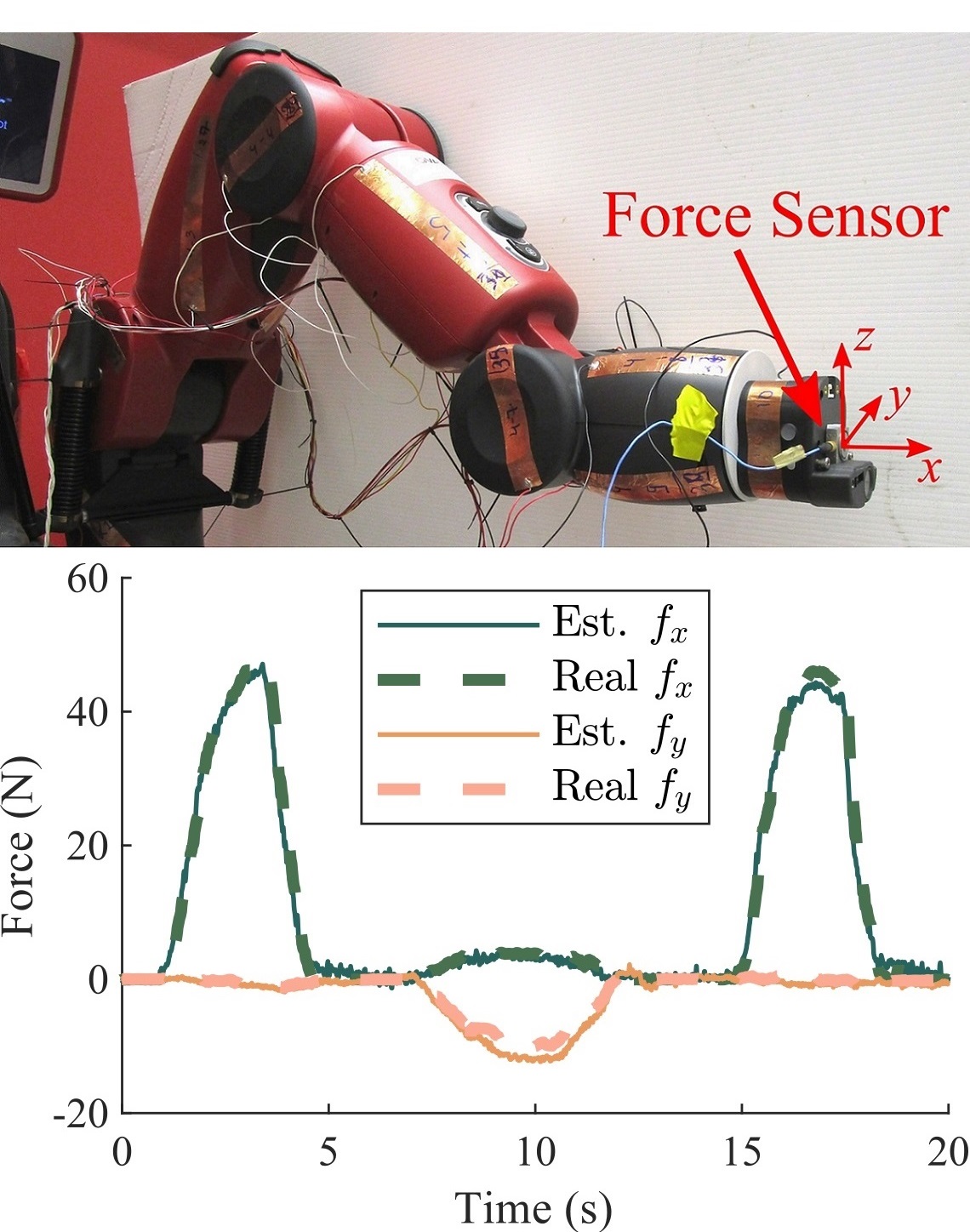}
	\label{fig:expFS-norm}}
\subfigure[]{
	\includegraphics[width = 0.225\textwidth]{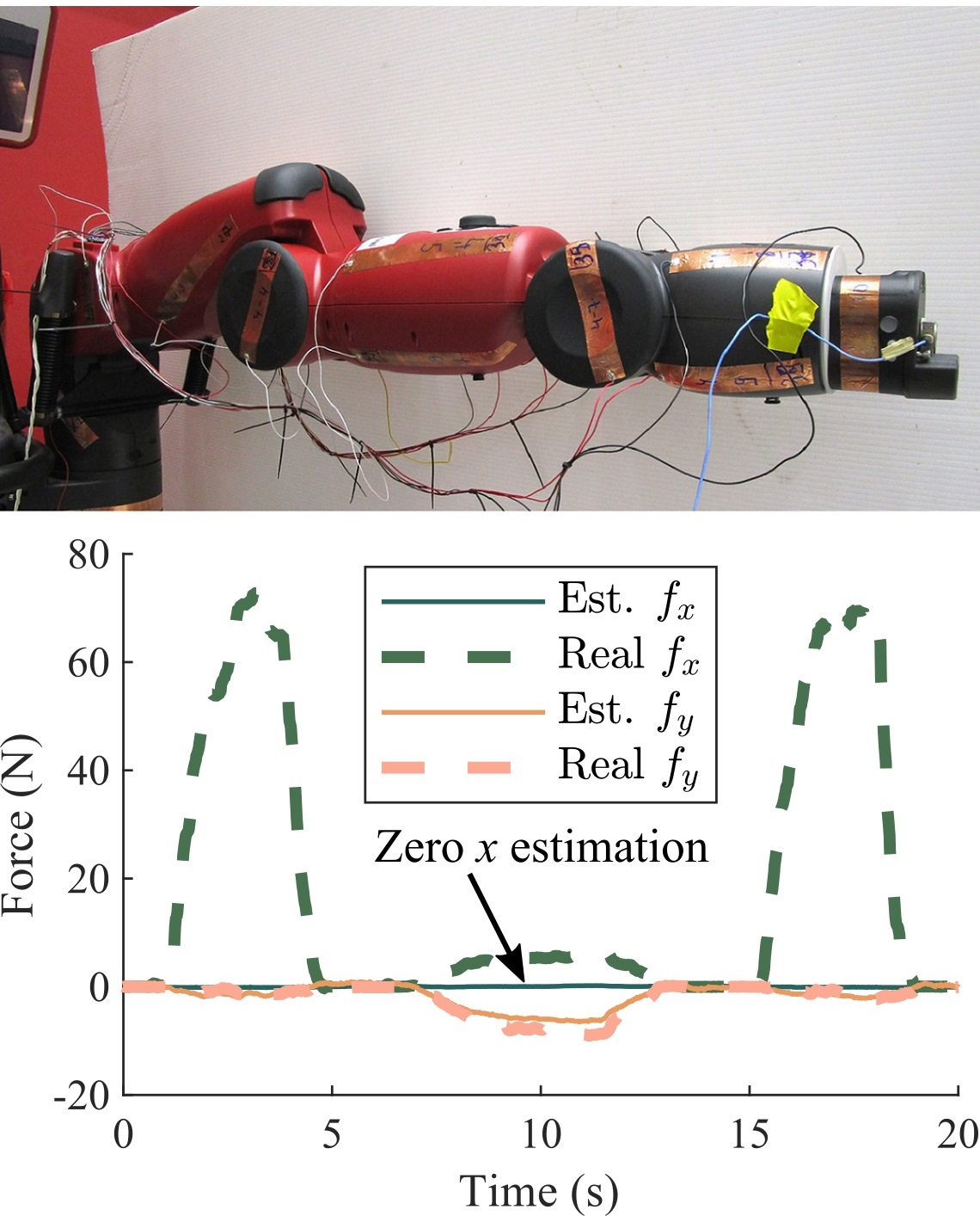}
	\label{fig:expFS-sing}}
\caption{Sensor observability experiments using the robot Baxter comparing EE forces estimated by the joint torque sensors and the ground truth as measured by an external 2-axis force sensor attached to the EE along the $x$- and $y$-axes.
The robot configurations include a) an arbitrary non-zero observability configuration and b) the observability singular configuration shown in Fig. \ref{fig:robotconfig-singsensX}. An external force is applied first in $x$, then $y$, and finally $x$ again.
Forces are fully observable in the arbitrary position in a), but forces in $x$ are not observable by the robot in the observability singular configuration in b).
}
\label{fig:expFS}
\end{figure}

Using the robot Baxter, we illustrate the importance of sensor observability index by observing changes in the robot's ability to use its joint sensors to estimate end effector forces in both normal and observability singular configurations.
End effector force estimation from joint sensors is performed using the packaged Baxter API from the manufacturer.
A 2-axis force sensor is attached onto the end effector of the robot as shown in Fig. \ref{fig:expFS-norm} to provide a ground truth for interaction forces along the $x$- and $y$-axes.
Once the robot is in position, the end effector is first pushed along the $x$-, then the $y$-, and finally $x$-axes again to observe whether the interaction forces are detected or not.

In the first scenario in Fig. \ref{fig:expFS-norm}, the robot is in an arbitrary non-zero observability configuration.
The associated plot shows the robot's ability to resolve the end effector forces using the joint torque sensors.
Conversely, in the second scenario, shown in Fig. \ref{fig:expFS-sing}, the robot is in the observability singular configuration shown in Fig. \ref{fig:robotconfig-singsensX}.
In this configuration, the robot's observability is zero in the $x$-axis and non-zero in $y$-axis.
As expected, from the plot, the robot is unable to observe interaction forces in the $x$-axis at $t \approx 3$ s and $t \approx 17$ s, despite the force sensor showing interaction forces, whereas $y$-axis forces are observed at $t \approx 10$ s. 
Off-axis forces are observed from imperfect interactions and the robot shifting during interaction.


%
%
%

\section{Conclusion} \label{sec:concl}


In this work, we introduce the novel concept of the sensor observability for analysing the quality of a specific joint configuration for observing task-space quantities.
We believe that this is the first work in quantifying the cumulating effect of distributed axial sensor positioning in multi-DOF robots and provides the base framework for developing further tools related to sensor analysis.
While it is most intuitively applied to force and torque sensing and the paper uses this case to demonstrate the application of sensor observability, the concept can be applied to other axial sensors such as accelerometers or distance sensors.
Future work, as mentioned above, will include further generalization of the concept to include unidirectional sensors, more formal comparisons with conventional methods and the use of the observability index in motion and robot design optimization.

\bibliographystyle{IEEEtran}
\bibliography{HumanoidReferences}

%

\end{document}